\documentclass[runningheads]{llncs}

\usepackage[T1]{fontenc}
\usepackage{graphicx}
\usepackage[table]{xcolor}
\usepackage{adjustbox}
\usepackage{amssymb}
\usepackage{amsmath}
\usepackage{stmaryrd}
\usepackage{dsfont}
\usepackage{subcaption}
\usepackage{pifont}
\usepackage{wrapfig}
\usepackage{adjustbox}
\usepackage{tabularx}
\usepackage{booktabs}
\usepackage{multirow}

\newcommand{\eg}{\textit{e}.\textit{g}., }
\newcommand{\q}[1]{`#1'}
\newcommand{\dq}[1]{``#1''}

\usepackage{environ}
\newcommand{\acksection}{\section*{Acknowledgments}} 
\NewEnviron{ack}{%
  \acksection
  \BODY
}

\begin{document}
\title{Concept Bottleneck with Visual Concept Filtering for Explainable Medical Image Classification}

\titlerunning{Concept Bottleneck with Visual Concept Filtering}
%
\newcommand*\samethanks[1][\value{footnote}]{\footnotemark[#1]}
\author{Injae Kim\thanks{Equal contribution}, Jongha Kim\samethanks, Joonmyung Choi, Hyunwoo J. Kim}
\authorrunning{Kim et al.}
%

\institute{Department of Computer Science and Engineering, Korea University \\
\email{\{dna9041, jonghakim, pizard, hyunwoojkim\}@korea.ac.kr}}
\maketitle              

\begin{abstract}
Interpretability is a crucial factor in building reliable models for various medical applications. Concept Bottleneck Models (CBMs) enable interpretable image classification by utilizing human-understandable concepts as intermediate targets. Unlike conventional methods that require extensive human labor to construct the concept set, recent works leveraging Large Language Models (LLMs) for generating concepts made automatic concept generation possible. However, those methods do not consider whether a concept is \textit{visually} relevant or not, which is an important factor in computing meaningful concept scores. Therefore, we propose a visual activation score that measures whether the concept contains visual cues or not, which can be easily computed with unlabeled image data. Computed visual activation scores are then used to filter out the less visible concepts, thus resulting in a final concept set with visually meaningful concepts. Our experimental results show that adopting the proposed visual activation score for concept filtering consistently boosts performance compared to the baseline. Moreover, qualitative analyses also validate that visually relevant concepts are successfully selected with the visual activation score.

\end{abstract}
\section{Introduction}
Deep Neural Networks (DNNs) have addressed many problems in various fields, including the medical domain~\cite{anwar2018medical,chang2018distributed,zhou2020diagnosis,xue2017preliminary}.
For instance,~\cite{zhou2020diagnosis} diagnoses breast lesions on dynamic contrast-enhanced MRI with deep learning, and~\cite{xue2017preliminary} diagnoses hip osteoarthritis using convolutional neural networks (CNNs). 
Despite the huge success of deep learning based models in the field of medical analysis and diagnosis, such models innately lack a crucial capability required in the medical domain - \emph{interpretability}.
Therefore, to tackle the difficulty in interpreting the model decision, a series of research have been presented to enhance the interpretability and the explainability of deep learning models.

Concept Bottleneck Models (CBMs)~\cite{koh2020concept} is one of the works that make image classification more interpretable.
Instead of directly predicting the target from non-interpretable latent representations, CBMs first predict concept scores, which measure the degree of an image matching to the human-understandable concepts. Then, the final prediction is done based on the predicted concept scores, which makes the prediction process interpretable.

Applying CBMs requires the construction of a set of concepts that well-describe images and are discriminable. Conventional approaches~\cite{koh2020concept,espinosa2022concept} try manually defining concepts, which requires extensive labor of skilled expert that is familiar with the target domain. However, such an approach largely hinders the scalability and generalizability of CBMs, since building large-scale concept set manually is costly, especially in the medical domain.

To address the issue, recent works~\cite{oikarinen2023label,yang2023language} propose generating concepts automatically by prompting Large Language Model (LLM) that contains rich information across various subjects to generate the concepts describing target classes.
Although such methods remove the need for manual concept generation, LLM turns out to generate non-visual concepts that do not align with images, therefore providing a noisy signal that hinders proper training.

To this end, we propose a \textit{visual activation score}, which measures whether a concept contains visual information or not.
By taking account of the visual activation score in the concept filtering phase via a submodular optimization~\cite{bach2010convex}, concepts that do not contain any visual cue useful for classifying the image are successfully removed from the initial concept set.

Experimental results demonstrate the effectiveness of the proposed visual activation score with consistent gains in terms of accuracy under multiple experimental settings.
Also, further analyses again validate that the visual activation score successfully discriminates the visual and non-visual concepts, therefore helping the model better classify the image with refined concepts.
\section{Related Works}
\subsection{Concept Bottleneck Models}
Concept Bottleneck Models (CBMs)~\cite{koh2020concept} aim to make the process of an image classification more interpretable, by implementing a concept bottleneck layer before the final classification layer. A concept bottleneck layer outputs a score of an image corresponding to multiple interpretable concepts.
Calculated concept scores are fed into a final linear layer to classify an image. Recent works~\cite{oikarinen2023label,yang2023language} propose leveraging information learned by Large Language Models (LLMs) to automatically extract concepts instead of manually constructing the concept set.

\subsection{Large Language Models}
Recently, Large Language Models (LLMs) based on Transformer~\cite{vaswani2017attention} architecture trained with the large-scale text corpus have shown to be effective in various downstream tasks including zero-shot and few-shot learning~\cite{brown2020language,touvron2023llama}, multi-task~\cite{radford2019language}, visual question answering (VQA)~\cite{touvron2023llama}. Moreover, leveraging knowledge of LLMs for computer vision tasks, such as generating candidate concepts to classify an image~\cite{oikarinen2023label,yang2023language}, is being studied recently.
\begin{figure}[t!]
    \centering
    \includegraphics[width=\linewidth]{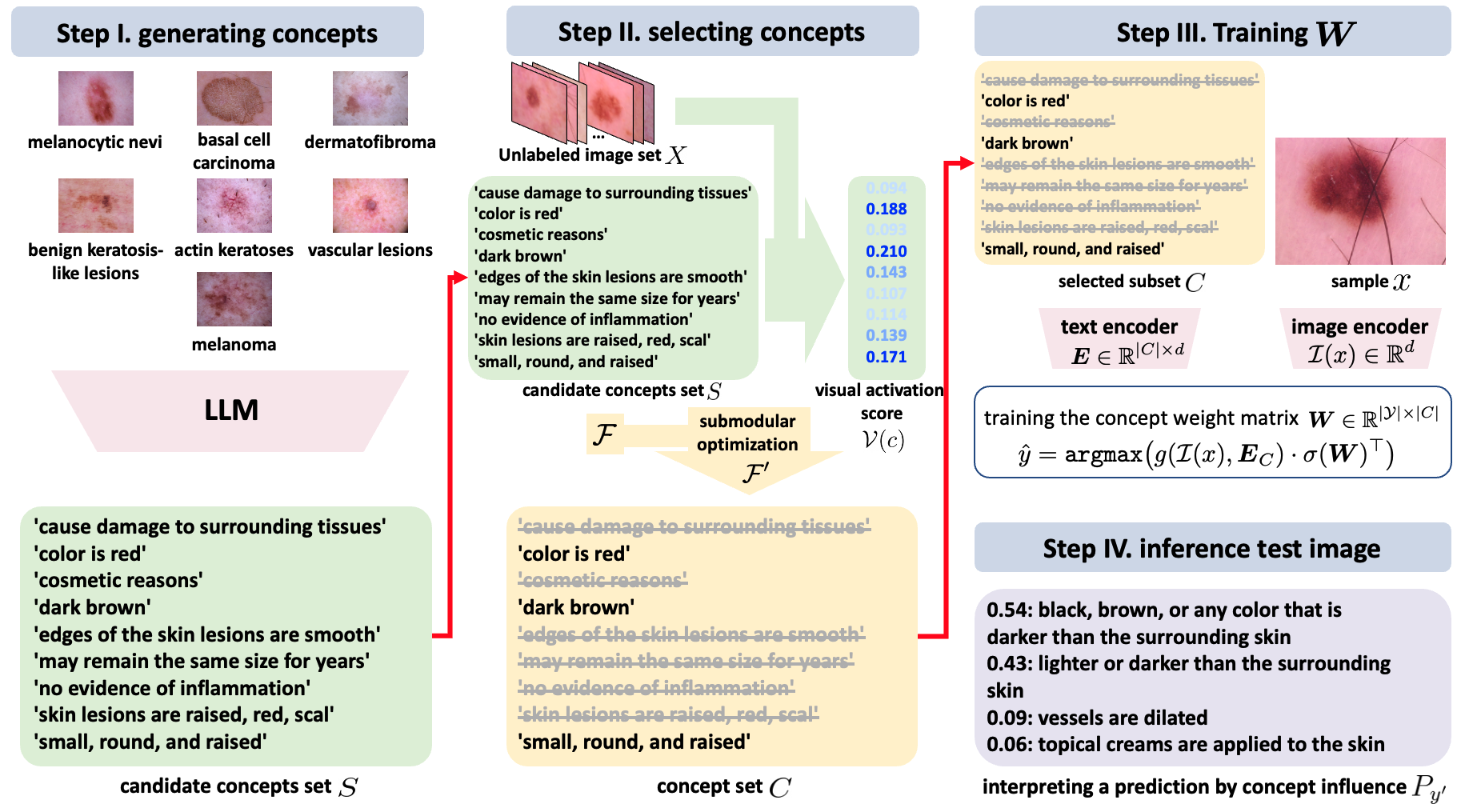}
    \caption{\textbf{Method Overview.} \textbf{Step 1}: 
    Generate candidate concepts set $S$ by prompting the large language model (LLM); \textbf{Step 2}: Select visually relevant concepts via submodular optimization with the score function $\mathcal{F}'$; \textbf{Step 3}: Train a concept weight matrix $W$ which projects concept scores into prediction logits; \textbf{Step 4}: Interpret inference results with the concept influence $P_{y'}$.}
    \label{fig:main_fig}
\end{figure}

\section{Method}
\subsection{Preliminary}
\label{sec:prelim}
Image classification is the task of predicting the class $y \in \mathcal{Y}$ an input image $x \in \mathcal{X}$ belongs to, where $\mathcal{Y}$ and $\mathcal{X}$ are sets of target classes and images, respectively.
In order to make the classification process more interpretable, Concept Bottleneck Models~\cite{koh2020concept,oikarinen2023label,yang2023language} firstly compute similarities between the image and a concept $c$ in the concept set $C$, which indicates how well the image and pre-defined concepts are aligned.
Computed similarities are then fed into the final classification layer to determine the class an image belongs to.
\paragraph{\textbf{Generating Candidate Concept Set}}
We adopt LaBo~\cite{yang2023language} as the baseline, which constructs the candidate concept set $S$ using GPT-3~\cite{brown2020language}. 
For each class $y$, LaBo prompts GPT-3 to retrieve sentences that describe the class.
500 sentences per class are retrieved from GPT-3 and split into shorter concepts to form a candidate concept set $S_y$ for each class.
Then, the whole concept set $S$ is defined as a union of $S_y$ as below:
\begin{equation}
\label{eq:union_concept}
    S = \underset{y\in \mathcal{Y}}{\cup}S_y
\end{equation}
\paragraph{\textbf{Concept Selection via Submodular Optimization}}
For a concept set $S_y$, submodular optimization~\cite{bach2010convex} is applied to select concepts with desired property using the score function $\mathcal{F}$, therefore resulting in a concept subset $C_y \subseteq S_y$, where $|C_y| = k$.
Due to the limitation in space, we refer to details about the submodular optimization to~\cite{yang2023language}.
The score function $\mathcal{F}$ that evaluates the utility of the subset $C_y$ is defined as:
\begin{equation}
\label{eq:submodular_function}
    \mathcal{F}(C_y) = \alpha \cdot \underset{c \in C_y}{\sum}D(c) + \beta \cdot \underset{c_1 \in S_y}{\sum} \underset{c_2 \in C_y} {\max} \phi(c_1, c_2),
\end{equation}
where $D(c)$ is a discriminability score of the concept $c$, $\phi(c_1, c_2)$ is a concept similarity between two concepts $c_1$ and $c_2$, and $\alpha$, $\beta$ are controllable hyperparameters. 
Maximizing the discriminability score $D(c)$ encourages selecting concepts that are aligned only with images with specific labels but not with the other images. 
To do so, the conditional likelihood $\overline{\text{sim}}(y|c)$ of a similarity score $\text{sim}(y,c)$ given a concept $c$ is defined as:
\begin{equation}
\overline{\text{sim}}(y|c) = \frac{\text{sim}(y,c)}{\sum_{y^\prime \in \mathcal{Y}}\text{sim}(y^\prime, c)}, \quad
\text{sim}(y,c)=\frac{1}{|\mathcal{X}_y|}\sum_{x\in \mathcal{X}_y} \mathcal{I}(x) \cdot \mathcal{T}(c)^\top,
\end{equation}
where $\mathcal{X}_y$ is training image set labeled with $y$, and $\mathcal{I}(\cdot)$ and $\mathcal{T}(\cdot)$ are the CLIP~\cite{radford2021learning} image encoder and text encoder, respectively.
Finally, $D(c)$ is defined as its negative entropy to maximize:
\begin{equation}
\label{eq:discriminability_score}
D(c) = \sum_{y^\prime \in \mathcal{Y}}\overline{\text{sim}}(y^\prime|c)\cdot \text{log}\Big(\overline{\text{sim}}(y^\prime|c)\Big).
\end{equation}
The second term of Equation~\ref{eq:submodular_function} is a coverage score that aims to maximize the minimum similarity between each concept in the subset $C_y$ and the overall set $S_y$.
With the coverage score, the selection of concepts covering a wide range of meanings of a target class is enabled.
Then, the whole concept set $C$ is defined as the union of $C_y$, analogous to the definition of $S$ in Equation~\ref{eq:union_concept}.
\paragraph{\textbf{Optimizing Concept Weight Matrix}}
After obtaining the concept set $C$, CLIP text features of concepts are stacked to form a concept embedding matrix $\boldsymbol{E}_C \in \mathbb{R}^{{|C|} \times d}$, where each row of $\boldsymbol{E}_C$ corresponds to a CLIP text embedding of a concept, $|C|$ is the size of the whole concept set $C$, and $d$ is a CLIP feature dimension.
With $\boldsymbol{E}_C$, a concept score $g(\mathcal{I}(x), \boldsymbol{E}_C) \in \mathbb{R}^{|C|}$ between $\boldsymbol{E}_C$ and an image $x \in \mathcal{X}$ is calculated as:
\begin{equation}
\label{eq:clip_scoring}
    g(\mathcal{I}(x), \boldsymbol{E}_C) = \mathcal{I}(x) \cdot \boldsymbol{E}_C^{\top}.
\end{equation}
Finally, a concept weight matrix $\boldsymbol{W} \in \mathbb{R}^{|\mathcal{Y}| \times |C|}$ which maps concept scores into the final prediction logit is optimized, where the final prediction $\hat{y}$ is computed as $\texttt{argmax}\big(g(\mathcal{I}(x), \boldsymbol{E}_C)\cdot \sigma(\boldsymbol{W})^\top\big)$.
$\sigma(\boldsymbol{W})$ denotes the softmax operation applied along the concept axis, where $\boldsymbol{W}_{y,c} = e^{\boldsymbol{W}_{y,c}} / \sum_{y' \in \mathcal{Y}} e^{\boldsymbol{W}_{y',c}}$.
As of the initialization,  $\boldsymbol{W}_{y,c}$ is set as 1 if $c \in C_y$ and 0 otherwise, in order to learn the weight $\boldsymbol{W}$ effectively in few-shot settings.
Given an image $x$, the concept influence $P_{y} \in \mathbb{R}^{|C|}$, which represents how much a concept influences the prediction of the class $y$ can be calculated as below:
\begin{equation}
P_{y} = g(\mathcal{I}(x), \boldsymbol{E}_c) \odot \sigma(\boldsymbol{W}_{y, *}),
\end{equation}
where $\odot$ is an element-wise multiplication operation.

\subsection{Concept Selection with Visual Activation Score}
In order to obtain a reliable concept score in Equation~\ref{eq:clip_scoring}, the concept $c \in C$ must include a \emph{visual cue}, e.g., \q{darker in color}.
However, the candidate concept set $S$ automatically extracted from LLM includes a lot of \emph{non-visual concepts}, which do not contain any visual cue, e.g., \q{most common type of precancerous lesion in the united states}. 
Those non-visual concepts hinder proper learning of the concept weight matrix $\boldsymbol{W}$ since scores of those concepts provide noisy signals.
Therefore, an appropriate criterion to filter out the non-visual concepts is required when constructing the concept subset $C$ via a submodular optimization. 

To measure the amount of visual information a concept contains, we define a scalar visual activation score $\mathcal{V}(c)$ of a concept $c$ defined as a standard deviation of CLIP scores between a concept $c$ and images $x \in X$ as below:
\begin{equation}
\label{eq:visual_activation_score}
    \mathcal{V}(c) = \text{stdev}(\{\mathcal{T}(c)\cdot\mathcal{I}(x)^{\top}\}_{x\in X}),
\end{equation}
where $X$ denotes an unlabeled target image set to calculate visual activation scores on, and $\mathcal{T}(c), \mathcal{I}(x) \in \mathbb{R}^d$ are the CLIP text embedding and the image embeddings of a concept $c$ and an image $x$, respectively.
As defined in the equation, the visual activation score of a concept $c$ is calculated as a standard deviation of concept scores among every image in the target image set $X$.
In other words, a concept that is activated differently depending on the image is regarded as a concept containing visual cues, since those concepts sensitively respond to visually distinct samples.
Note that an arbitrary dataset can be set as $X$ since it does not require labels.
Further analyses regarding the utilization of various datasets as X are provided  in Section~\ref{sec:analysis_dataset}.
Calculated $\mathcal{V}(c)$ is then added to the original score function $\mathcal{F}$ to form a new score function  $\mathcal{F}'$ as follows:
\begin{equation}
    \mathcal{F'}(C_y) = \alpha \cdot \underset{c \in C_y}{\sum}D(c) + \beta \cdot \underset{c_1 \in S_y}{\sum} \underset{c_2 \in C_y} {\max} \phi(c_1, c_2) + \gamma \cdot \underset{c \in C_y}{\sum}\mathcal{V}(c),
\end{equation}
With the new score function $\mathcal{F}'$, the subset $C_{y'}$ is obtained via a submodular optimization, and the following procedures are done analogously as described in Section~\ref{sec:prelim}.

\section{Experiments}
We validate the effectiveness of the proposed method by applying the method to HAM-10000~\cite{tschandl2018ham10000}, a skin disease dataset.
The dataset consists of 10,015 dermatoscopic images collected from various patients. 
The target classes $\mathcal{Y}$ consist of 7 types of skin problems, Melanocytic Nevi, Benign Keratosis-like Lesions, Dermatofibroma, Vascular Lesions, Actinic Keratoses, Basal Cell Carcinoma, and Melanoma.
For all experiments, we follow the experimental settings of LaBo~\cite{yang2023language} except for the hyperparameters $\alpha, \beta, \text{and}~\gamma$.
All experiments are done with a single NVIDIA RTX A6000 GPU.

{\renewcommand{\arraystretch}{1.0}%
\begin{table}[t!]
    \caption{\textbf{Performance on HAM-10000 dataset.} \dq{Number of shots} denotes the number of labeled samples per target class, where \q{Full} denotes that the model is trained with the whole dataset. * denotes reproduced results.
    }
    \centering
    
    \setlength{\tabcolsep}{3.5pt}  
    
    \begin{adjustbox}{width=0.95\textwidth}
    \begin{tabular}{l|cccccc}
        \toprule
         \multirow{2}{*}{Method} & \multicolumn{6}{c}{Number of Shots} \\
         \cline{2-7}
         & 1 & 2 & 4 & 8 & 16 & Full\\
        \midrule
        \textcolor{gray}{Linear Probe$^*$} & \textcolor{gray}{44.4} & \textcolor{gray}{58.5} & \textcolor{gray}{44.9} & \textcolor{gray}{49.0} & \textcolor{gray}{61.5} & \textcolor{gray}{82.5} \\
        \midrule
        LaBo\cite{yang2023language}$^*$ & 36.5 & 44.9 & 44.5 & 43.0 & 58.5 & 80.8  \\
        LaBo\cite{yang2023language}$^*$ + Ours & 53.2 \small \textcolor{blue}{(+16.7)} & 45.4 \small \textcolor{blue}{(+0.5)} & 47.4 \small \textcolor{blue}{(+2.9)} & 46.1 \small \textcolor{blue}{(+3.1)} & 61.4 \small \textcolor{blue}{(+2.9)} & 81.0\small \textcolor{blue}{(+0.2)}\\
        \bottomrule
    \end{tabular}
    \end{adjustbox}
    \label{tab:main_table}
\end{table}
}

    
    
\subsection{Experimental Results}
In Table~\ref{tab:main_table}, a consistent gain in accuracy under every single setting compared to LaBo is reported by applying the visual activation score.
In terms of filtering out non-visual concepts, LaBo's discriminability score $D(c)$ also indirectly refines non-visual concepts by maximizing the negative entropy of $\overline{\text{sim}}(y|c)$. 
However, the fewer label data, the more inaccurate value of $\overline{\text{sim}}(y|c)$ is, and at 1-shot, the LaBo achieved a low accuracy of 36.7\%.
By using an unlabeled image set $X$, the visual activation score $\mathcal{V}(c)$ encourages to filter non-visual concepts effectively, therefore outperforming LaBo by 16.5\% at 1-shot.
Also, while maintaining the interpretability, in some cases, the proposed method is shown to even outperform performances of linear probing where the classification process is completely non-interpretable.

\begin{figure}[t!]
    \centering
    \includegraphics[width=\linewidth]{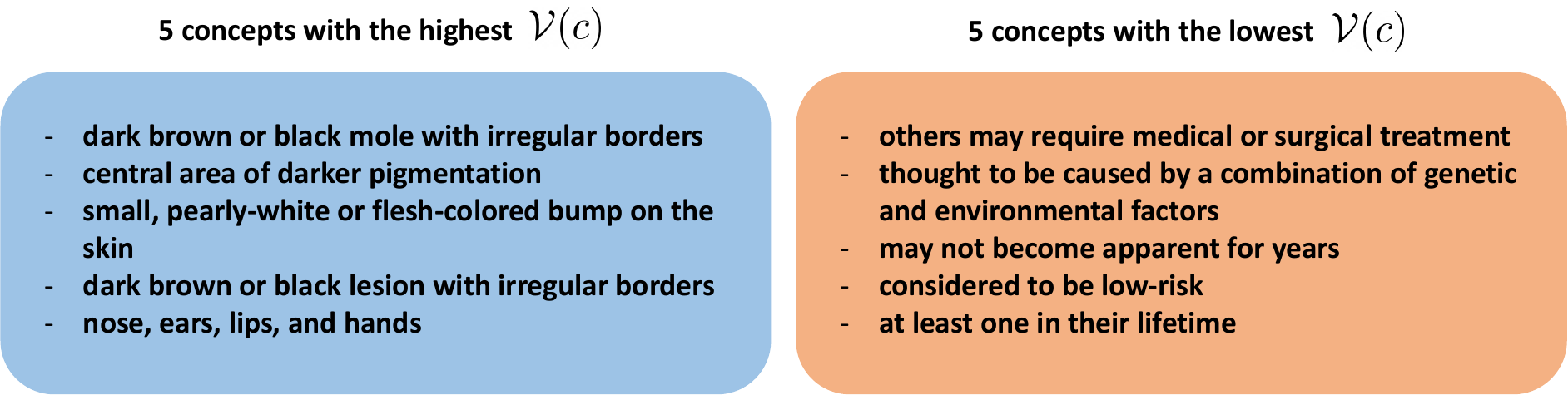}
    \caption{
    Examples of concepts that have the highest and the lowest visual activation score $\mathcal{V}(c)$ are listed.
}
    \label{fig:Vcfigure}
    \vspace{-0.5cm}
\end{figure}

\section{Analysis}
\subsection{Analysis on a Visual Activation Score $\mathcal{V}(c)$}
In Figure~\ref{fig:Vcfigure}, five concepts with the highest and the lowest visual activation score $\mathcal{V}(c)$ are listed, respectively.
As shown in the figure, concepts with high $\mathcal{V}(c)$ turn out to be visual concepts, such as \dq{dark brown or black mole with irregular borders}.
On the other hand, concepts with low $\mathcal{V}(c)$ are mostly non-visual or meaningless concepts, including \dq{others may require medical or surgical treatment}. 
Such qualitative examples demonstrate that the visual activation score $\mathcal{V}(c)$ acts as an effective measure to detect non-visual or meaningless concepts.

{\renewcommand{\arraystretch}{1.0}%
\begin{table}[t!]
    \caption{\textbf{Analysis on the image set $X$ for visual activation score.} \dq{Number of shots} denotes the number of labeled samples per target class, where \q{Full} denotes that the model is trained with the whole dataset.
    }
    \centering
    
    \setlength{\tabcolsep}{3.5pt}  
    
    \begin{adjustbox}{width=0.95\textwidth}
    \begin{tabular}{l|cccccc}
        \toprule
         \multirow{2}{*}{Image set $X$} & \multicolumn{6}{c}{Number of Shots} \\
         \cline{2-7}
         & 1 & 2 & 4 & 8 & 16 & Full\\
        \midrule
        w/o $\mathcal{V}(c)$& 36.5 & 44.9 & 44.5 & 43.0 & 58.5 & 80.8  \\
        \midrule
        HAM10000 & \textbf{53.2} & 45.4 & 47.4 & 46.1 & 61.4 & \textbf{81.0}  \\
        ImageNet & 48.5 & \textbf{46.2} & \textbf{50.3} & \textbf{56.8} & 62.1 & 80.9  \\
        COCO & ~~~47.7~~~ & ~~~45.4~~~ & ~~~48.5~~~ & ~~~45.7~~~ & ~~~\textbf{62.3}~~~ & ~~~80.8~~~  \\

        \bottomrule
    \end{tabular}
    \end{adjustbox}
    \label{tab:ablation_image_set}
\end{table}
}
\subsection{Analysis on Target Image Set $X$}
\label{sec:analysis_dataset}
In Table~\ref{tab:ablation_image_set}, experimental results with multiple target dataset $X$ to calculate the visual activation score $\mathcal{V}(c)$ in Equation~\ref{eq:visual_activation_score} are provided. 
We conduct experiments under adopting train splits of HAM-10000~\cite{tschandl2018ham10000}, ImageNet~\cite{deng2009imagenet}, and COCO~\cite{lin2014microsoft} as $X$.
For the ImageNet and COCO datasets, 10,000 images are randomly sampled to match the size of $X$ to that of the HAM-10000 dataset.
The result shows that overall performance gains are reported regardless of the target set $X$. 
Such results show the effectiveness of the proposed method in that it does not require a domain-specific dataset, but instead can be implemented using an arbitrary dataset.
We anticipate that such property could facilitate the application of the proposed method to diverse medical domains, where acquiring large-scale domain-specific target images is expensive.

\begin{figure}[t!]
    \centering
    \includegraphics[width=\linewidth]{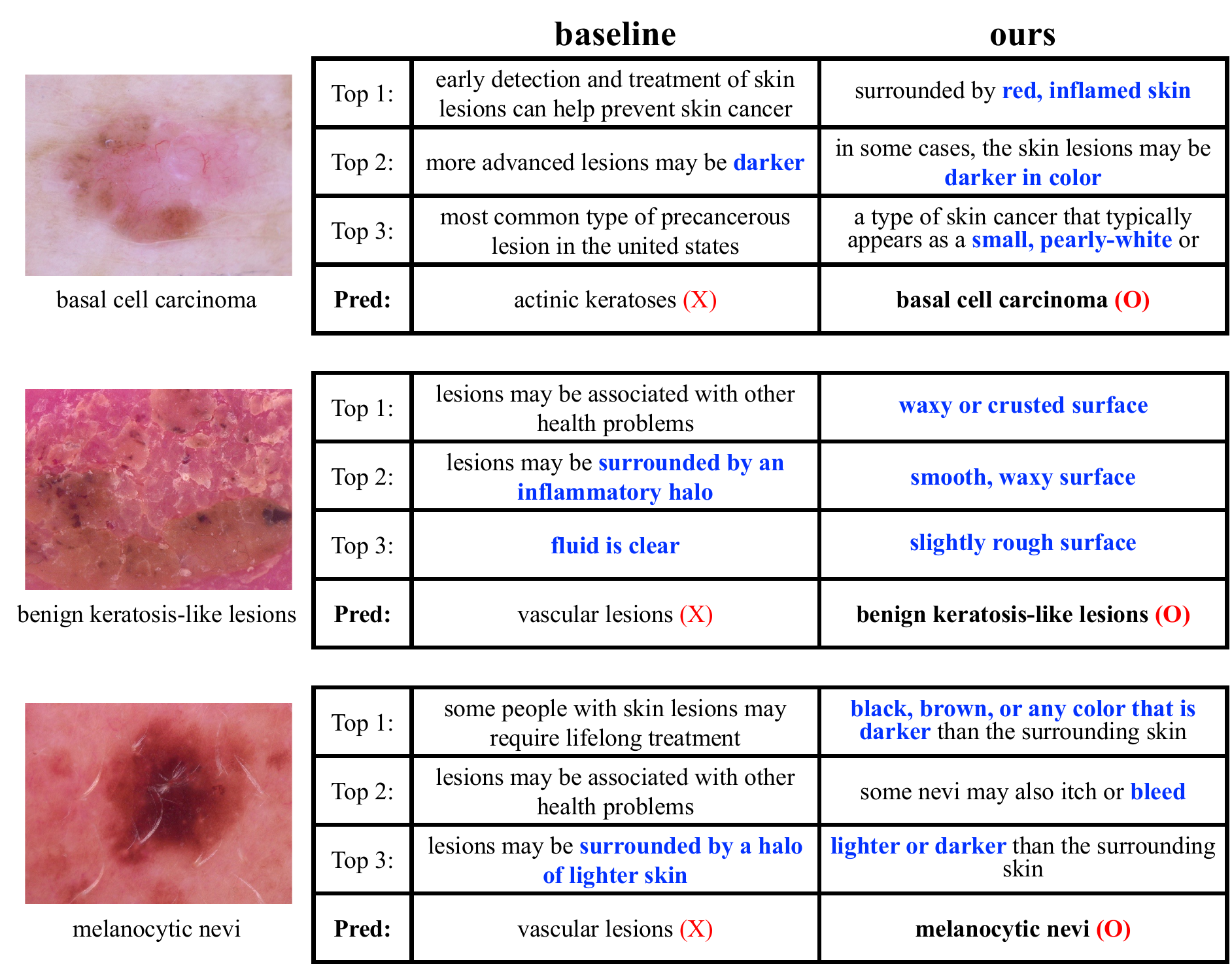}
    \caption{\textbf{Qualitative results on HAM-10000 dataset.} Several comparisons with the baseline, LaBo, are shown with the top-3 concepts which are ranked by their weights in the linear function. Parts that provide visual cues are colored \textbf{\textcolor{blue}{blue}}.
}
    \label{fig:qualitative}
\end{figure}

\subsection{Qualitative Examples}
In Figure~\ref{fig:qualitative}, the final prediction of a model and three concepts with the highest influence $P_y$ in predicting the ground-truth class $y$ are illustrated.
As depicted in the figure, the baseline fails correctly classifying the image since the concepts with the highest influences in predicting the ground-truth class are mostly non-visual concepts \eg \dq{early detection and treatment of skin
lesions can help prevent skin cancer}, \dq{lesions may be associated with other
health problems} which does not help classify an image. 
In contrast, the prediction result on the same image is corrected when applying the proposed method since visually irrelevant concepts are removed from $S_y$, thus concepts with the highest influences are replaced with concepts with visual cues.
The result validates that adding the visual activation score $\mathcal{V}(c)$ to the score function $\mathcal{F}$ helps filter visually irrelevant concepts, therefore contributing to better classification results by providing concepts that are rich in information.
\section{Conclusion}
In this paper, we propose a method to refine the non-visual concepts generated from large language models (LLMs) which hinder the training of Concept Bottleneck Models (CBMs). 
In order to filter out non-visual concepts, we propose the visual activation score which measures whether a concept contains visual information or not. 
With computed visual activation scores of concepts, non-visual concepts are filtered out via a submodular optimization.
Quantitative and qualitative analyses demonstrate that the proposed visual activation score contributes to detecting and filtering out non-visual concepts, therefore resulting in consistent improvement in accuracy.

\begin{ack}
This research was supported by the MSIT (Ministry of Science and ICT), Korea, under the ICT Creative Consilience program (IITP-2023-2020-0-01819) supervised by the IITP(Institute for Information \& communications Technology Planning \& Evaluation).
\end{ack}


%
%
%
%




\bibliographystyle{unsrt}
{\small
\bibliography{main}
}
\end{document}